%% file: main.tex
\documentclass[lettersize,journal]{IEEEtran}

\input{Sections/packages}

\begin{document}

\title{Scalable Decentralized Cooperative Platoon using Multi-Agent Deep Reinforcement Learning}

\author{Ahmed Abdelrahman$^{*1}$, Omar M. Shehata$^{1}$, Yarah Basyoni$^{1}$, and Elsayed I. Morgan$^{1}$

\thanks{$^{1}$Faculty of Engineering and Material Science, German University in Cairo (GUC), Cairo, Egypt\\
$^{*}$Corresponding Author: ahmad.salah@ieee.org}

}



\maketitle

\begin{abstract}
Cooperative autonomous driving plays a pivotal role in improving road capacity and safety within intelligent transportation systems, particularly through the deployment of autonomous vehicles on urban streets. By enabling vehicle-to-vehicle communication, these systems expand the vehicles' environmental awareness, allowing them to detect hidden obstacles and thereby enhancing safety and reducing crash rates compared to human drivers who rely solely on visual perception. A key application of this technology is vehicle platooning, where connected vehicles drive in a coordinated formation. This paper introduces a vehicle platooning approach designed to enhance traffic flow and safety. Developed using deep reinforcement learning in the Unity 3D game engine, known for its advanced physics, this approach aims for a high-fidelity physical simulation that closely mirrors real-world conditions. The proposed platooning model focuses on scalability, decentralization, and fostering positive cooperation through the introduced predecessor-follower 'sharing and caring' communication framework. The study demonstrates how these elements collectively enhance autonomous driving performance and robustness, both for individual vehicles and for the platoon as a whole, in an urban setting. This results in improved road safety and reduced traffic congestion.

\end{abstract}

\begin{IEEEkeywords}
Cooperative Autonomous Driving, Vehicle Platooning, Deep Reinforcement Learning.
\end{IEEEkeywords}

\section*{Supplementary Videos}
This paper is accompanied by a narrated video of the performance of the developed scalable decentralized cooperative platoon: https://youtu.be/nktS4ctIj9A

\input{Sections/sec1_Introduction}
\input{Sections/sec3_Methodology}
\input{Sections/sec4_ExperimentalWork}

\input{Sections/sec5_Results}

\input{Sections/sec6_conclusion}
\input{Sections/ref}

\end{document}

%% file: Sections/packages.tex
\usepackage{stfloats}
\usepackage{url}
\usepackage{verbatim}
\usepackage{cite}
\usepackage{algorithm} 
\usepackage{algorithmic}  
\usepackage[algo2e]{algorithm2e} 
\usepackage{amsmath,amssymb,amsfonts}
\usepackage{algorithm}
\usepackage{graphicx}
\usepackage{textcomp}
\def\BibTeX{{\rm B\kern-.05em{\sc i\kern-.025em b}\kern-.08em
    T\kern-.1667em\lower.7ex\hbox{E}\kern-.125emX}}
    
    \usepackage{lineno}
\usepackage{lettrine}

\usepackage{url}
\usepackage{moreverb}

\usepackage{graphicx}
\usepackage{subcaption}
\usepackage{float}
\usepackage{amsmath}
\usepackage{array}
\usepackage{multirow}
\graphicspath{ {Figures/} }

\usepackage{scalerel}
\usepackage{tikz}
\usetikzlibrary{svg.path}

\definecolor{orcidlogocol}{HTML}{A6CE39}
\tikzset{
  orcidlogo/.pic={
    \fill[orcidlogocol] svg{M256,128c0,70.7-57.3,128-128,128C57.3,256,0,198.7,0,128C0,57.3,57.3,0,128,0C198.7,0,256,57.3,256,128z};
    \fill[white] svg{M86.3,186.2H70.9V79.1h15.4v48.4V186.2z}
                 svg{M108.9,79.1h41.6c39.6,0,57,28.3,57,53.6c0,27.5-21.5,53.6-56.8,53.6h-41.8V79.1z M124.3,172.4h24.5c34.9,0,42.9-26.5,42.9-39.7c0-21.5-13.7-39.7-43.7-39.7h-23.7V172.4z}
                 svg{M88.7,56.8c0,5.5-4.5,10.1-10.1,10.1c-5.6,0-10.1-4.6-10.1-10.1c0-5.6,4.5-10.1,10.1-10.1C84.2,46.7,88.7,51.3,88.7,56.8z};
  }
}

\newcommand\orcidicon[1]{\href{https://orcid.org/#1}{\mbox{\scalerel*{
\begin{tikzpicture}[yscale=-1,transform shape]
\pic{orcidlogo};
\end{tikzpicture}
}{|}}}}

\usepackage[hidelinks,bookmarks=false]{hyperref} 

%% file: Sections/sec1_Introduction.tex
\section{Introduction}
\IEEEPARstart{E}{ra} of automation is rising and Autonomous Vehicles (AVs) shape and provide an important technology that contributes to creating this era. AVs, as one of the automation applications that many companies and research communities are hitting, are coming to our streets to assist and improve the safety standards of intelligent transportation systems (ITS). Levels of AVs range from level 1, just as driver assistant, to level 5, fully automated driving vehicle, as provided in J3016 standard \cite{TaxonomyAD} that illustrated the classification levels of autonomous driving. Sensors are what shape a vehicle's perception and let vehicles observe the surrounding environment to know how to properly interact with it, but each sensor like camera and radar has its field of view (FOV) that limits the vehicle's perception, as also for human driver's vision. \\
Driver error is the main cause of vehicle accidents and driver inattention is one of the highest causes of these accidents which yields to approximately 80\% of vehicle crashes and 65\% of near-crashes as reported in the conducted study by the National Highway Traffic Safety Administration \cite{dingus2006100}.
Cooperative autonomous driving tackles such a serious cause of vehicle accidents by providing a wider perception of AVs. Vehicles can communicate with each other through vehicle-to-vehicle (V2V) to alert surrounding vehicles by their position that may be for other vehicles without communication are out of their FOV. Intersections with buildings limit vehicles' observation to the coming crossing cars and based on the report by the Fatality Analysis Reporting System (FARS) and National Automotive Sampling System General Estimates System (NASS-GES) data, approximately 40\% of the estimated 5,811,000 crashes that happened in the US in 2008 were intersection-related crashes \cite{choi2010crash}. Numbers and statistics force increasing intentions to invest more work and time in ITS applications and technologies. \\
One of these well-known applications for cooperative autonomous driving is vehicle platooning, which is a group of connected AVs driving with certain formations following each other. Vehicle platooning has a quite high impact on enhancing safety, increasing highway capacity, and decreasing traffic congestion\cite{kavathekar2011vehicle}. V2V communication between vehicles in the platoon enables cooperative control strategies that can achieve high Cooperative ITS (C-ITS) objectives through coordination across all vehicles in the platoon. There are many developed communication topologies for vehicle platooning as predecessor follower, bidirectional predecessor follower, two-predecessors following, and others \cite{li2017distributed}, \cite{zheng2015stability}. These different types of inter-communication between vehicles aim to achieve better cooperation for the whole system performance and how vehicles can create cooperative driving strategies to improve the impact of C-ITS on cities' roads. However, some of these communication topologies require too much processing power and high communication capabilities like 5G or better to handle the high flow of data between vehicles, so big data issues should be taken into consideration to avoid any lag or miscommunication between the connected vehicles.\\
Two of the main objectives of vehicles in the platoon are decreasing gap errors between the vehicles and maintaining the same velocity between them throughout the whole driving distance. Researchers studied the performance of classical approaches for vehicle platooning control such as backstepping, sliding mode control, and model predictive control (MPC) \cite{shalaby2019design} to manage the gaps between vehicles in platoon formation, and others addressed the usage of artificial intelligence such as deep reinforcement learning (RL) \cite{chu2019model}. Classical approaches for vehicle platooning control achieved good results, however, some limitations exist in these types of controllers, such as the need to develop a very highly accurate vehicle dynamical model, which increases the complexity of the system as the model dynamics complexity increases, and the need to high computational power for the nonlinear optimization. Farag et al. used a deep RL-based controller and MPC for managing the inter-vehicle distances in a platoon, and the analysis of the controllers confirms that the RL controller outperforms the MPC in terms of computational time and control effort while maintaining a similar root mean square error in the inter-vehicle distances \cite{faragreinforcement}. Also, many other studies demonstrated the power of model-free deep RL for vehicle motion control \cite{jaritz2018end}, \cite{cai2020high}, and some of the work exceeded superhuman driving performance \cite{fuchs2021super}.

After examining the above-mentioned literature in addition to the survey conducted on applying deep RL for autonomous vehicles’ control \cite{aradi2020survey}, there exist some research gaps listed as follows:

\begin{itemize}
    \item Most of the research papers and theses apply the dynamical vehicle model in studying a small number of vehicles, and for a large number of vehicles, they use the kinematical model or a simplified linearized dynamical model.
    \item The use of low physics simulators, where there is a huge gap between the simulation and real-life work.
    \item Urban test case experiments are not addressed heavily for vehicle platooning, for the complexity of the urban environment and the presence of surrounding vehicles affect the ego vehicle’s control actions.
\end{itemize}

The contribution of this paper is to develop a scalable decentralized cooperative control hierarchy using Deep RL that optimizes the performance of platoon behavior and cooperation starting from a platoon of a single vehicle to multiple vehicles without affecting computational expenses. Simulation work is held on Unity, a very powerful real-time 3D physics Game Engine. An urban environment is created on Unity and different testing scenarios are conducted to evaluate the proposed multi-agent system (platoon) performance and robustness as illustrated in the following sections.\\
The rest of the paper is organized as follows; Section II introduces the proposed platoon methodology, and then in Section III, the experimental work of this work is illustrated and clarifies how results in Section IV are presented, while Section V concludes all the work of the paper.

%% file: Sections/sec3_Methodology.tex
\section{Methodology}
This section presents the procedures for developing a Scalable Decentralized Cooperative Platoon (SDCP). Connected autonomous vehicles are trained together to learn how to drive on a track with turns and obstacles while learning how to drive in a platoon formation and respect its regulations like keeping the same gap between vehicles and the same velocity as the predecessor vehicle. Figure 1 shows the proposed Markov Decision Process (MDP) for this multi-agent reinforcement learning problem. MDP framework supports formalizing the loop between the agents and the environment \cite{littman1994markov}; agents interact with the environment by taking actions that result in a new state with feedback from the environment in the shape of rewards that gives a clue to the learning behavior's action whether it is a good or bad decision. SDCP should be able to maximize the performance of this cooperative task, platoon formation, and each agent/vehicle's single performance. The environment is developed to help agents learn this behavior to be greedy for maximizing rewards from solo driving and cooperative driving.

\begin{figure}[H]
    \centering
    \includegraphics[scale=0.35]{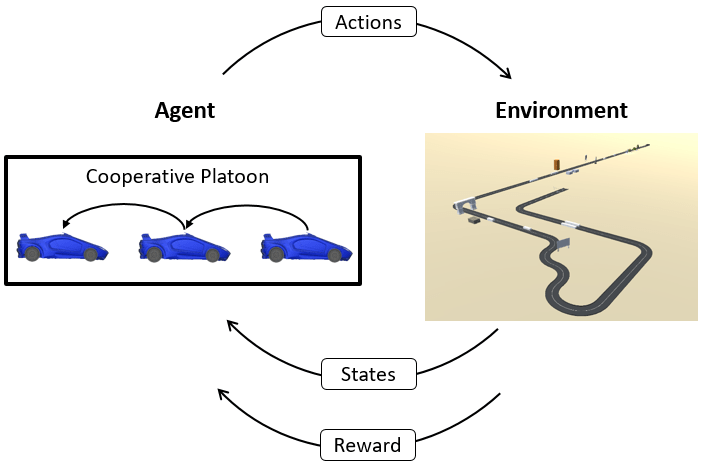}
    \caption{MDP for SDCP}
\end{figure}

\subsection{Environment}
The work is conducted in an urban environment with certain rules, such as vehicles driving on the right lane, no crashes, and car-following driving behavior. Track structure, checkpoints, and obstacles form the environment, as shown in Figure 2, which the agents will interact with. Environment corresponds to agents' actions by sending back positive or negative rewards based on the taken action and presenting new situations to the agents. With these interactions, vehicles learn to drive autonomously while maintaining safe driving and respecting platoon formation.

\begin{figure}[H]
    \centering
    \includegraphics[scale=0.35]{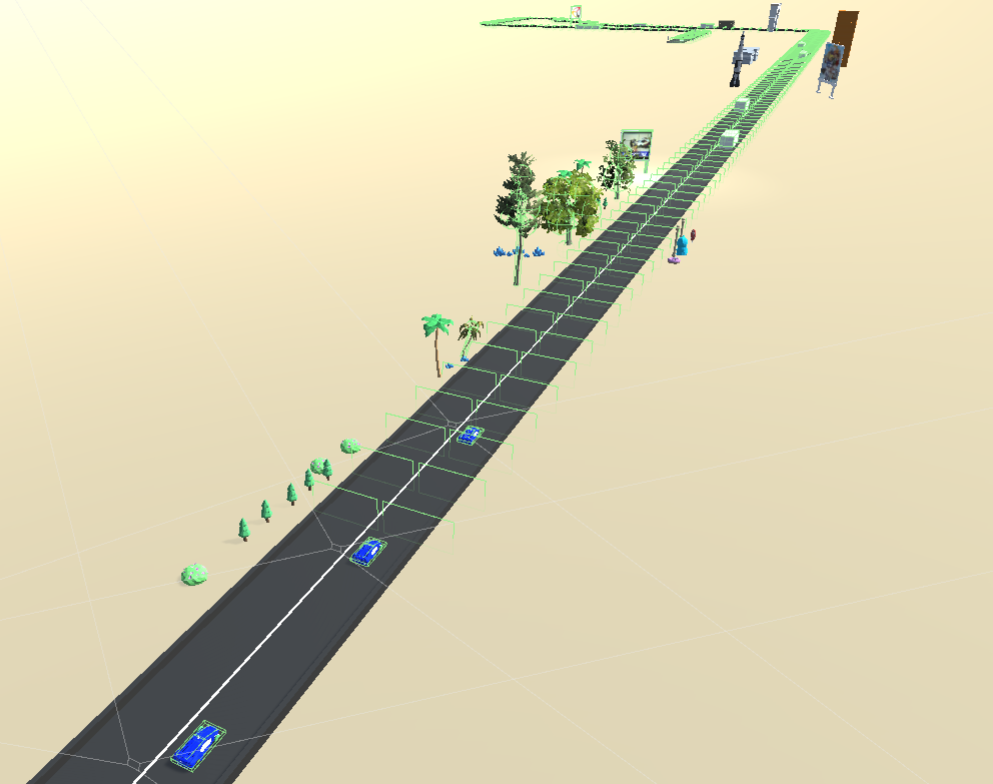}
    \caption{Training Environment}
    \label{fig:my_label}
\end{figure}

\subsubsection{Track Design}
The track is designed to scale up the complexity of driving scenarios. First, vehicles learn to drive on a straight long highway road on the right lane while avoiding hitting road borders or other vehicles, then complexity increases by avoiding crashing static obstacles, and then by driving through complex turns.

\subsubsection{Checkpoints}
Right and left lanes are full of checkpoints, invisible non-rigid colliders, as shown in Figure 3, which guide vehicles through rewards to learn that they should drive in the right lane unless to avoid crashing, which will result in higher negative reward than to do lane change and drive on the left lane.

\begin{figure}[H]
    \centering
    \includegraphics[scale=0.35]{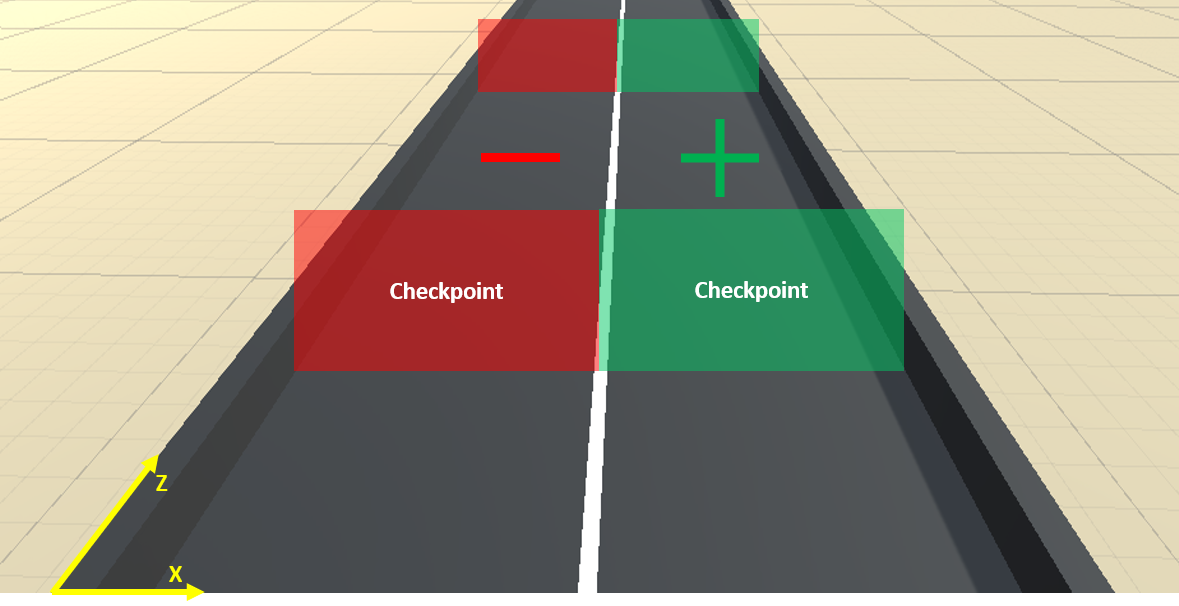}
    \caption{Checkpoints}
\end{figure}

\subsubsection{Obstacles}
There are two types of obstacles. Static obstacles, which are rigid solid boxes, represent a scenario of a stopped vehicle, closed road, or an obstacle that would force vehicles to do lane change.
Dynamic obstacles are the other vehicles in the platoon.

\subsection{Multi-Agent}
Three vehicles are connected and trained using model-free Deep RL, where model-free RL is a powerful general tool for learning complex behaviors \cite{pong2018temporal}, to learn how to drive on the track while focusing on safe driving, by avoiding crashing with track borders, static obstacles, or dynamic obstacles (other vehicles).
The developed vehicle dynamical model by Unity \cite{UnitySA} is used in this work to have a highly accurate and realistic dynamic model that quite matches real-life physics.

\begin{figure}[H]
    \centering
    \includegraphics[scale=0.375]{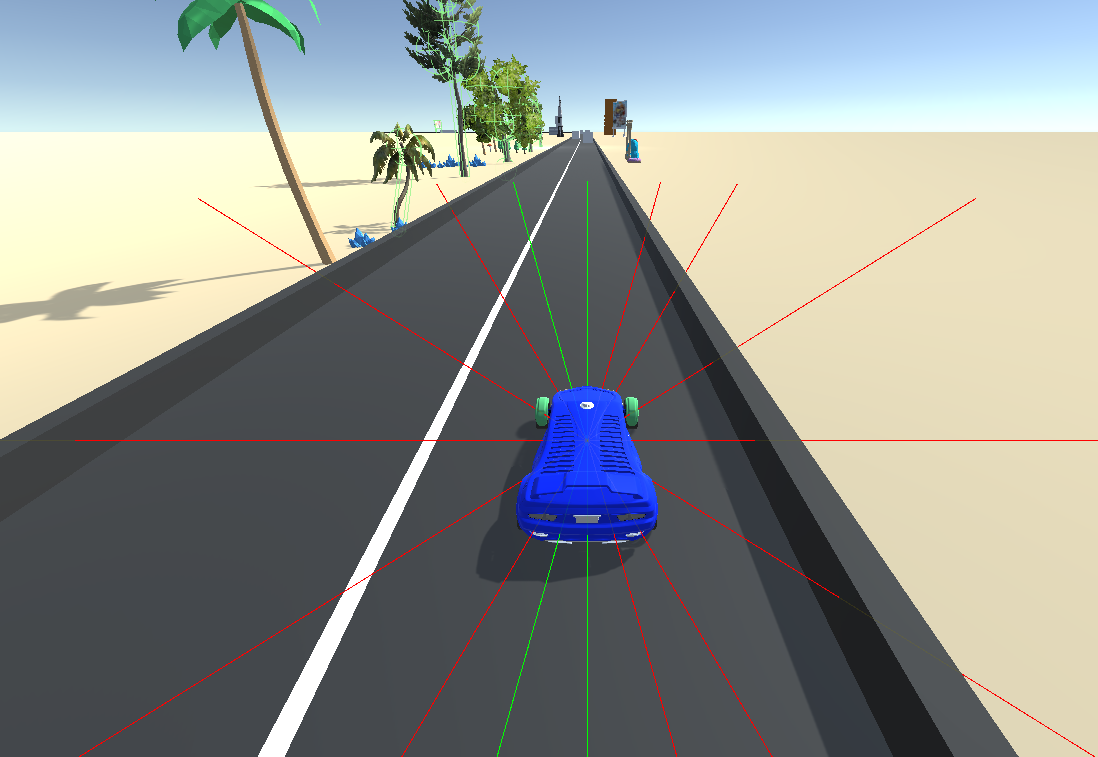}
    \caption{Agent/Vehicle}
    \label{fig:my_label}
\end{figure}

\textbf{Ego Perception}
The vehicle needs to observe the surrounding environment to learn how to drive. These observations are also the inputs to the deep neural network of the training model. The vehicle's perception consists of 16 raycast sensors, in which each raycast is like a depth sensor that measures the distance to the facing objects within its range of measuring as visualized in Figure 4, the vehicle's velocity, the gap between it and the predecessor vehicle and the shared data through V2V communication.

\subsection{Scalable Decentralized Cooperative Platoon (SDCP)}
SDCP's foundational principles include V2V communication, decentralization, and scalability. V2V communication forms the backbone of the cooperation, facilitating the seamless exchange of data between the AVs, which is essential for coordinating movements and ensuring safety. Decentralization plays a crucial role, as it allows each vehicle to operate autonomously and make decisions independently, enhancing the system's resilience and flexibility. This decentralization is crucial in scenarios where the lead vehicle may encounter issues, allowing for quick reorganization without compromising the platoon's functionality. Lastly, scalability is a key principle, ensuring that the system can efficiently adapt to varying sizes of the vehicle platoon without a significant impact on performance or safety. This scalability is vital for accommodating different traffic conditions and platoon configurations, making the SDCP a robust solution for a wide range of real-world applications.\\
In practice, both leader and follower vehicles operate without prior knowledge of road layouts or obstacles, focusing on safe navigation. Safety is the paramount concern, even more so than maintaining platoon formation. For example, if the platoon leader is involved in an accident, the system adapts by reassigning the lead role to the first follower, ensuring both continuity and safety. The leader vehicle prioritizes safe driving, not collecting perception data from followers but rather providing its own sensory information to them. This approach maintains formation and avoids risky maneuvers or actions that might lead to accidents or disrupt the platoon's cohesion, demonstrating the SDCP's commitment to safety and efficiency in various real-world scenarios.\\
The below subsections illustrate SDCP core concepts:\\

\subsubsection{Sharing and Caring Communication Topology}
V2V communication utilizes a predecessor-follower topology, chosen for its minimal impact on computational load as the number of vehicles in a platoon increases. Although alternate topologies may offer superior control, the predecessor-follower system is being refined to enhance platoon scalability and high driving performance without encountering big data challenges \cite{zheng2014influence}. This refinement is influenced by the "sharing and caring" concept, as originated and described by this study \cite{abdelrahman2023development}, which boosts the cooperation between connected AVs. The "sharing and caring" concept is set to be integrated into the predecessor-follower communication topology to foster cohesive cooperation among connected autonomous vehicles.
Through this improved protocol, AVs not only focus on their performance but also consider the welfare and performance of others in the platoon. This enhances cooperation as vehicles share key data like raycast sensor readings and velocity. Importantly, before executing any critical maneuvers, vehicles assess the potential impact on the platoon, especially avoiding actions that could lead to accidents. This cooperative approach promotes safer autonomous driving, strengthens the bond between connected vehicles, and increases the collective rewards for all vehicles in the platoon. Figures 5 and 6 in the referenced document visually depict these differences in communication topology among platoon vehicles.

\begin{figure}[H]
    \centering
    \includegraphics[scale=0.5]{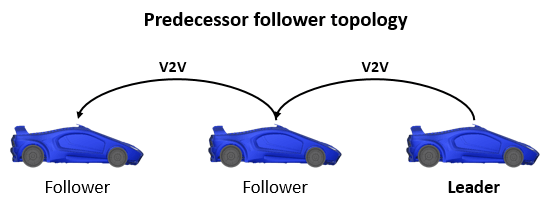}
    \caption{Predecessor Follower Communication Topology}
    \label{fig:my_label}
\end{figure}

\begin{figure}[H]
    \centering
    \includegraphics[scale=0.5]{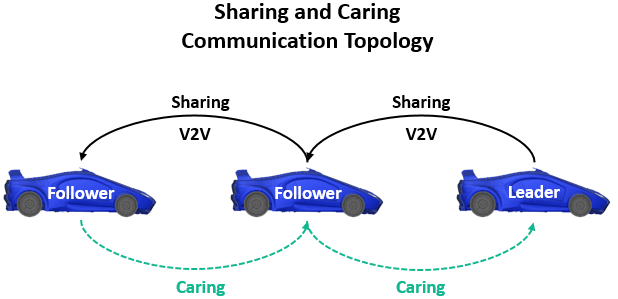}
    \caption{Predecessor Follower-based Sharing and Caring Communication Topology}
    \label{fig:my_label}
\end{figure}

\subsubsection{Decentralized Training}
Vehicles are trained together while sharing their data and caring for each other's cumulative rewards based on the developed communication topology, however, each agent is also greedy for its cumulative rewards.
Forcing each agent to focus on maximizing both its rewards and cooperation rewards enables the ability to have both dependent and independent behaviour. The vehicle will depend on other vehicles to gain more rewards from cooperation, but if this cooperation will cause large negative rewards to the agent, it will decide to be independent and focus on driving safely till reaching the desired destination. So for a platoon as an application or form of cooperative driving, vehicles will learn how to drive in a group and respect the platoon formation while being able to drive safely and solely if needed.

\subsubsection{Scalability}
The above-mentioned concepts (sharing and caring communication topology and decentralization) made it applicable to have a scalable number of vehicles in a platoon. It is like a chain where platoon length can be increased or decreased starting from a platoon consisting of a single vehicle to n-number of vehicles as long as this length can fit into the road design and structure.

\subsection{RL Reward Signals}
Reward shaping is a very critical aspect of RL training performance as it is the way of communication to the agent to evaluate it in order to achieve the required behavior.
Curriculum Learning\cite{bengio2009curriculum} is applied on penalties like those caused by crashes to let vehicles have the courage to train with fewer worries about negative rewards at the beginning of training and then penalties' values are increased to give higher priority to safe driving and avoid any crashes or platoon dissociation. If penalties are high from the start of training, sometimes agents decide to stop driving or are afraid of taking any action that may lead to high negative rewards and this slows down the training progress and may lead to passive, not active behavior.

All vehicles in the platoon are subjected to the same positive rewards as shown in Table I, which are the reward signals that focus on teaching vehicles that they should drive in the right lane and as fast as possible to collect more positive rewards. Penalties or negative rewards are the reward signals that teach vehicles safe driving (avoid crashing with track borders, obstacles, or other vehicles). These numerical values for reward signals are subjected based on a trial and error method.
Table I presents rewards for only the discrete events, while speed and gap error penalties are subjected to each time step.

\begin{table}[H]
\centering
\caption{Reward Signals for Discrete Events} 
\resizebox{8.5cm}{!}{
\renewcommand{\arraystretch}{1.5}
\begin{tabular} {|c||c|c|c|} 
 \hline
 \textbf{Event} & \textbf{Reward [0-10m]} & \textbf{Reward [10-15m]} & \textbf{Reward [15-25m]}\\
 \hline
 Checkpoints of right lane & +1 & +1 & +1\\
 \hline
 Checkpoints of left lane & -0.1 & -1 & -2\\
\hline
Reach finish line & +100 & +100 & +100\\
\hline
Follower vehicle reaches finish line & +50 (caring) & +50 (caring) & +50 (caring)\\
\hline
Crash & -10 (no caring) & -10 (no caring) & -50 (no caring)\\
\hline
\end{tabular} }
\end{table}

To apply caring between agents, agents should learn that working cooperatively results in positive rewards, and not helping each other results in negative rewards. Each follower vehicle in the platoon cares about the predecessor vehicle and this is applied by giving a positive reward to the predecessor vehicle if the follower reaches the finish line and a negative reward in case of crashes. This trains the predecessor vehicle to avoid any aggressive actions that may lead to crashes or disorder behavior among the platoon.

The formulated penalty equation of velocity error for the leader:
\begin{gather}
e_{velocity,l} = V_l - V_{desired} \\
P_{velocity,l} = -(e_{velocity,l}/V_{max})^2
\end{gather}  

where $e_{velocity,l}$ is the calculated error between the desired velocity for the platoon and leader vehicle’s velocity, and $P_{velocity,l}$ is the corresponding applied velocity penalty on the leader per stamp time.

Followers are subjected to penalties for gap and velocity errors:

\begin{gather}
e_{velocity,f} = V_f - V_{predecessor} \\
P_{velocity,f} = -(e_{velocity,f}/V_{max})^2\\
e_{gap,f} = G_f - G_{desired}\\
P_{gap,f} = -(e_{gap,f}/G_{max})^2
\end{gather}

where $e_{velocity,f}$ is the velocity error between the follower and predecessor vehicles, $P_{velocity,f}$ is the corresponding applied velocity penalty on the follower vehicle, $e_{gap,f}$ is the measured error in the gap between the follower and predecessor vehicle and $P_{gap,f}$ is the subjected penalty on the follower vehicle for the gap.
The gap ($G_f$) between the ego vehicle’s position ($x_ego$, $z_ego$) and the predecessor vehicle’s position ($x_{predecessor}$, $z_{predecessor}$) is calculated as follows:

where the gap between the ego vehicle and the predecessor vehicle is calculated as follows:
\begin{gather}
G_f = \sqrt{(x_{ego} - x_{pred}) + (z_{ego} - z_{pred})}
\end{gather}

Velocity and gap error equations are inspired by the work of this study \cite{faragreinforcement} which succeeded in formalizing equations that ease learning for agents in vehicle platooning formation. The vehicle's top speed is 40km/h, the desired gap between vehicles is 10 meters, and the maximum allowable gap between vehicles is 30 meters. The gap error penalty ranges from 0 (10m) to -1 (30m) while if the gap increases to be more than 30m, then the gap penalty is the same as 30m of -1. Rewards on gap and speed errors are applied on each time step of 0.02 seconds. However, other rewards are applied when an event happens like crashing or reaching the finish line.

%% file: Sections/sec4_ExperimentalWork.tex
\section{Experimental work}

\subsection{Setup}
The hardware setup that hosts the training and testing experiments is a personal computer with Intel Core i7-9750H CPU, Nvidia GeForce RTX 2060 GPU, and 16GB of RAM. Unity real-time 3D Game Engine\cite{Unity} is the simulator adopted for creating the training environment and evaluating the proposed platoon structure with the help of the developed ML-Agents toolkit\cite{juliani2018unity} within Unity that enables the integration between Unity Editor and training the multi-agent cooperative platoon.

\subsection{Training}
A platoon of only three vehicles is trained using Proximal Policy Optimization (PPO) based on end-to-end deep reinforcement learning to learn a robust cooperative driving behavior. PPO has shown better results over other RL algorithms while being more empirical for having better sample complexity \cite{juliani2018unity},\cite{schulman2017proximal}, \cite{ye2020automated}. Most of the PPO hyperparameters are set to the default values supported by the ML-Agents toolkit or adopted from other studies' recommendations as in \cite{gupta2021embodied}. Deep Neural Network consists of 5 layers of shape \{35,256,256,256,2\}. Figure 7 visualizes the Network, the input layer that takes ego perception, the gap between follower and predecessor vehicles, and shared predecessor perception. Then 3 fully connected hidden layers with 256 neurons each, and the output layer consists of two neurons which are the steering and throttle of each vehicle. All vehicles in the platoon have the same trained model with the same neural network layer, except for the leader vehicle the V2V perception data input is off.

\begin{figure}[H]
    \centering
    \includegraphics[width=0.7\linewidth]{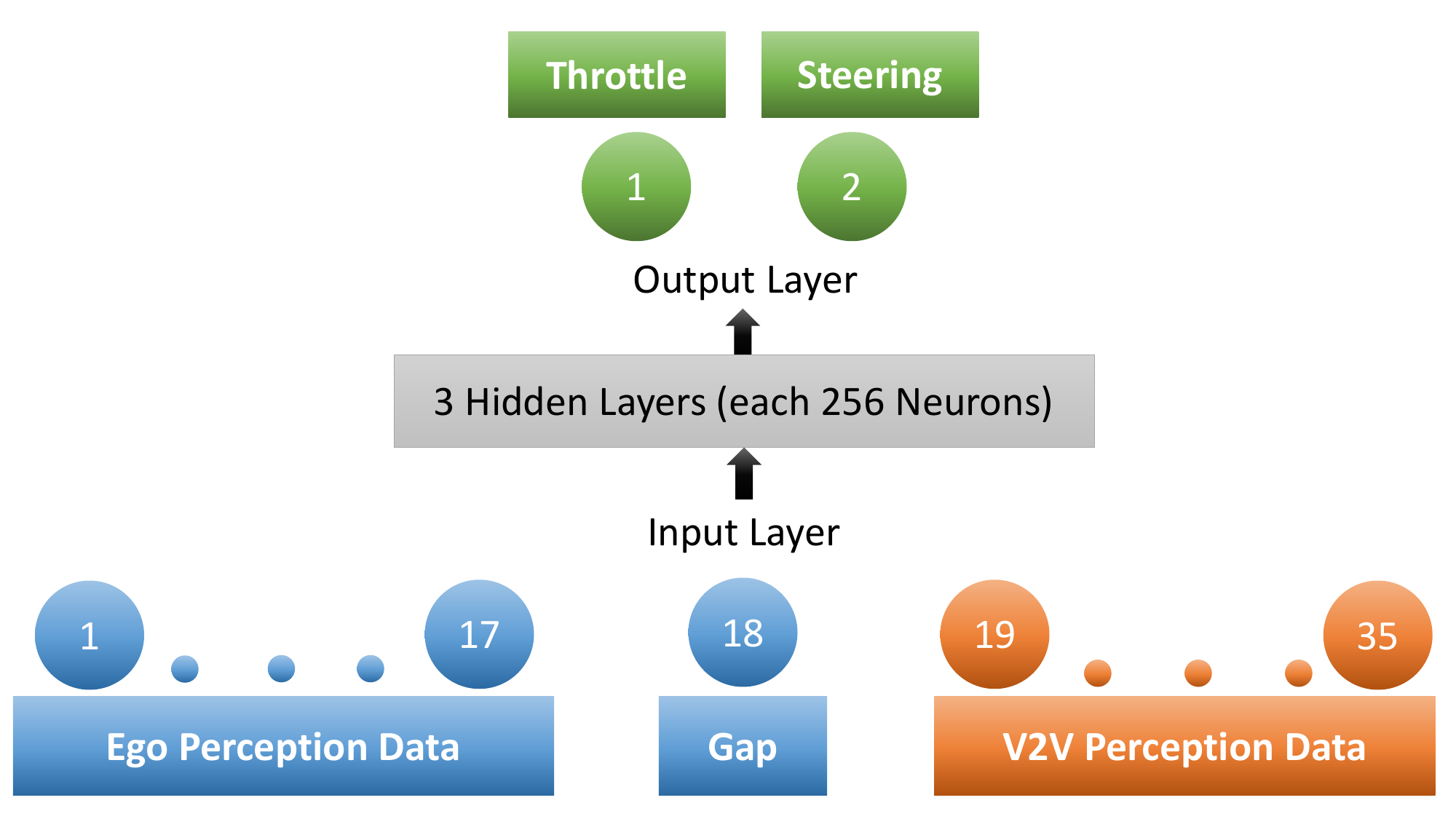}
    \caption{Deep Neural Network Layers}
    \label{fig:my_label}
\end{figure}

Figure 8, shows the training track with a total driving distance of 1300 meters. Vehicles are trained to drive safely in a cooperative system (platoon). while caring about their rewards and other vehicles' rewards like avoiding penalties for crashing. Training is applied on two parallel environments with the same track design but the positions of obstacles are changed to avoid model over-fitting and parallel environments decrease the training time.
\begin{figure}
    \centering
    \includegraphics[width=9cm,height=5cm,keepaspectratio]{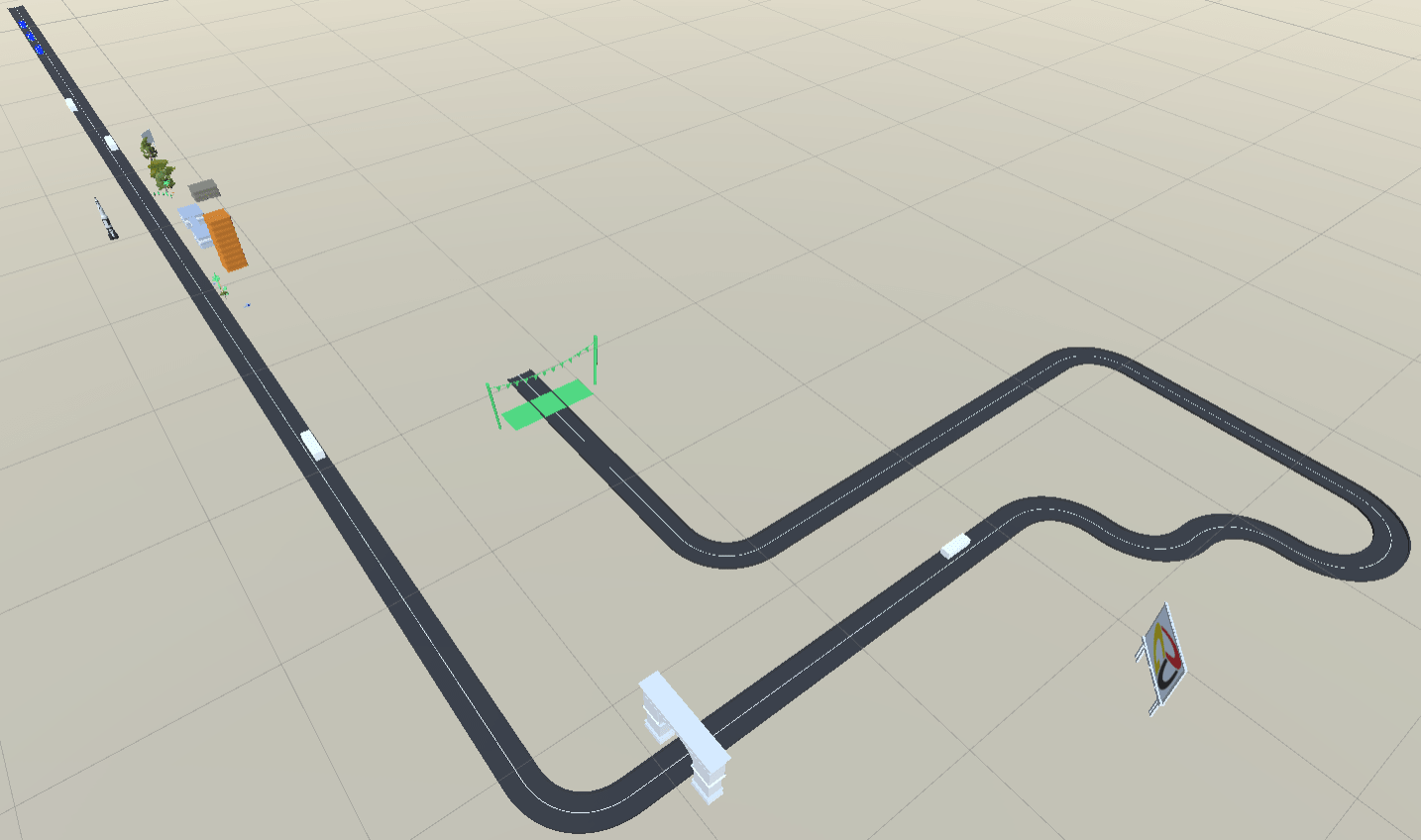}
    \caption{Urban Training Track}
    \label{fig:my_label}
\end{figure}

\subsection{Evaluation metrics}
 SDCP performance will be evaluated by Root Mean Square Error (RMSE), Standard Deviation (Std), maximum gap, and stalled gap values for the gap error between the vehicles and the number of crashes which is one of the main perspectives of safe driving validation.
The test track is designed to be as complex as the training track but with new scenarios like passing through a narrow way between two obstacles as shown in Figure 9 and this complexity is to simulate real hard maneuvers for the human drivers themselves. Agents have no prior knowledge of the test track design or obstacle locations and their objective is to keep driving in platoon formation and respect its rules as much as possible while being able to handle driving safely and solely if needed. Table II shows the driving distance for each test case experiment.\\
Five test case scenarios are conducted to test the performance of SDCP and each test case is to validate one of the core concepts of the developed platoon. The first test case is to test scalability by increasing the number of vehicles in the platoon from three vehicles, as in the training process, to eight vehicles. The second test is for measuring the cooperation between vehicles in dealing with aggressive maneuvers from the leader vehicle. The third test is for validating decentralization where followers are subjected to drive without the leader, and this simulates a real-life scenario like if the leader vehicle of the platoon has a malfunction and the platoon should continue without the leader till reaching the desired destination for time-saving. The fourth test is for the whole platoon, SDCP, driving on the testing track. The fifth test case is for measuring the impact of the developed SDCP on decreasing traffic congestion.

\begin{table}
\centering
\caption{Driving distance for each test case} 
\resizebox{8.5cm}{!}{
\renewcommand{\arraystretch}{2}
\begin{tabular} {|c||c|c|c|c|c|c|} 
 \hline
 \textbf{Test Case no.} & \textbf{Test Case 1} & \textbf{Test Case 2} & \textbf{Test Case 3} & \textbf{Test Case 4} & \textbf{Test Case 5}\\
 \hline
 Track length & 750 m & 750 m & 750 m & 1000 m & 1000 m\\
 \hline
\end{tabular} }
\end{table}

\begin{figure}
    \centering
    \includegraphics[scale=0.3]{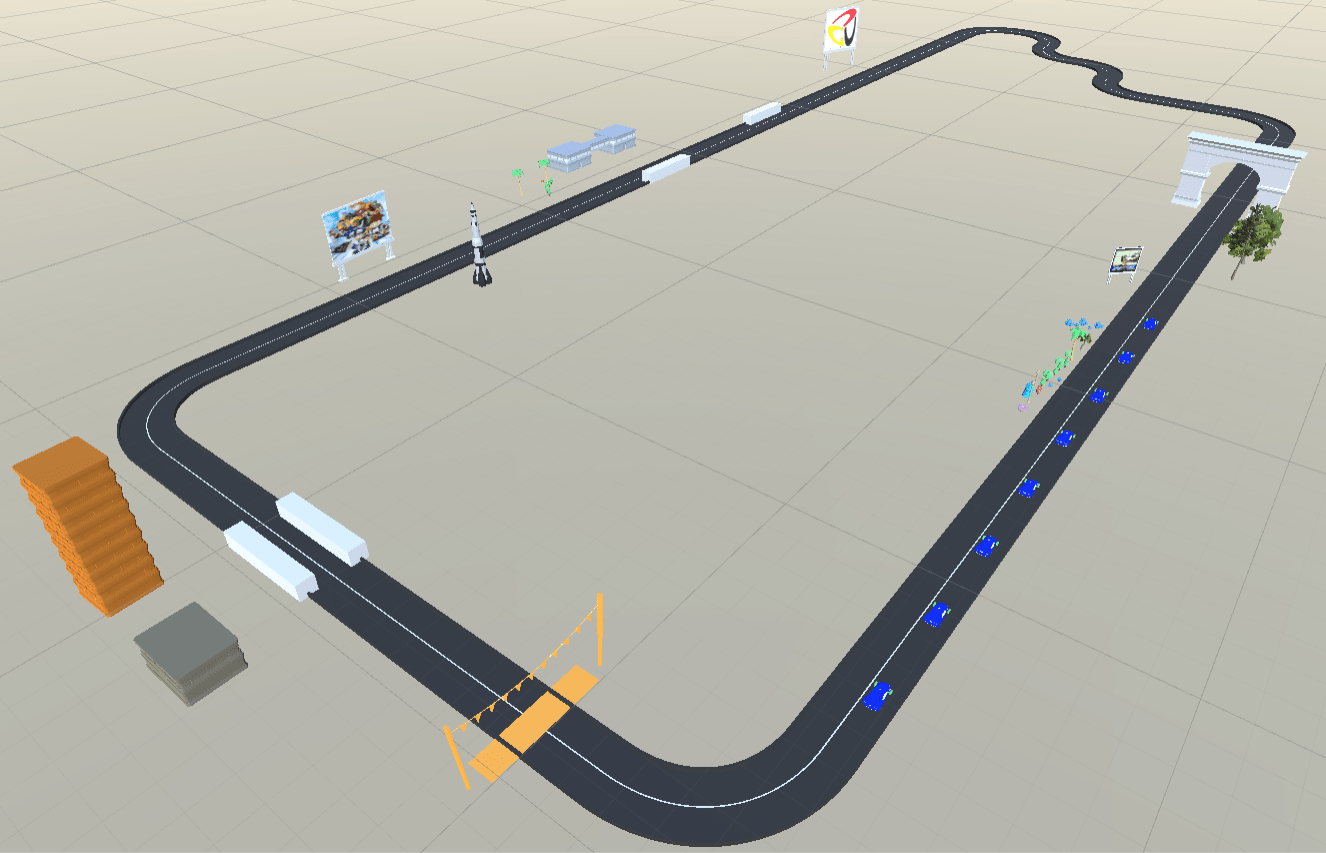}
    \caption{Urban Testing Track}
    \label{fig:my_label}
\end{figure}

%% file: Sections/sec5_Results.tex
\section{Results and Discussion}
In this section, the results of SDCP performance will be presented along with test scenarios introduced in section III and provide more details in this section. The discussion highlights the advantages and limitations of the proposed platoon system.

The training model for this multi-agent system (SDCP) took 25 million training steps for approximately 11.25 hours with a timescale x10, which is 112.5 hours of training on the real-time scale. Figure 10 shows the mean cumulative reward of the training model for each 20000 training step and the curve is smoothed to make a better visualization of learning mean rewards between exploration and exploitation.

\begin{figure}
    \centering
    \includegraphics[height=4cm, width=9cm]{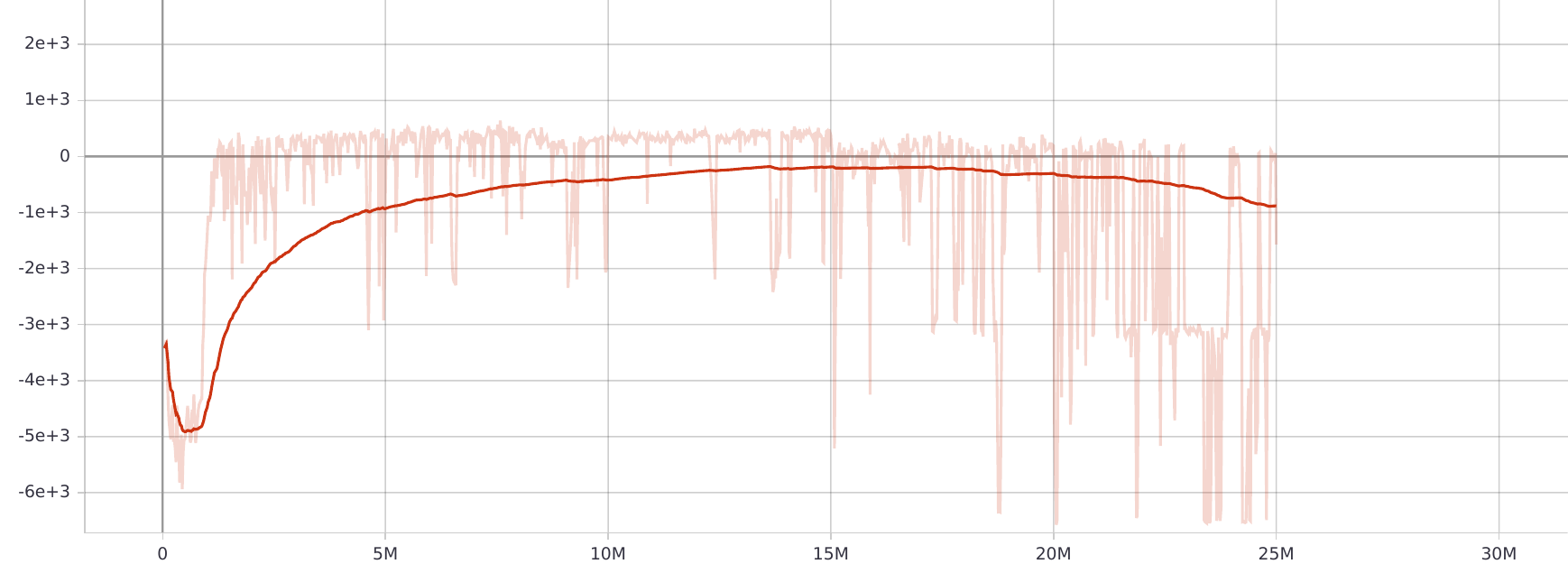}
    \caption{Mean Cumulative Reward vs Training Steps for the Multi-Agent System}
\end{figure}

\begin{figure}
    \centering
    \includegraphics[width=1\linewidth]{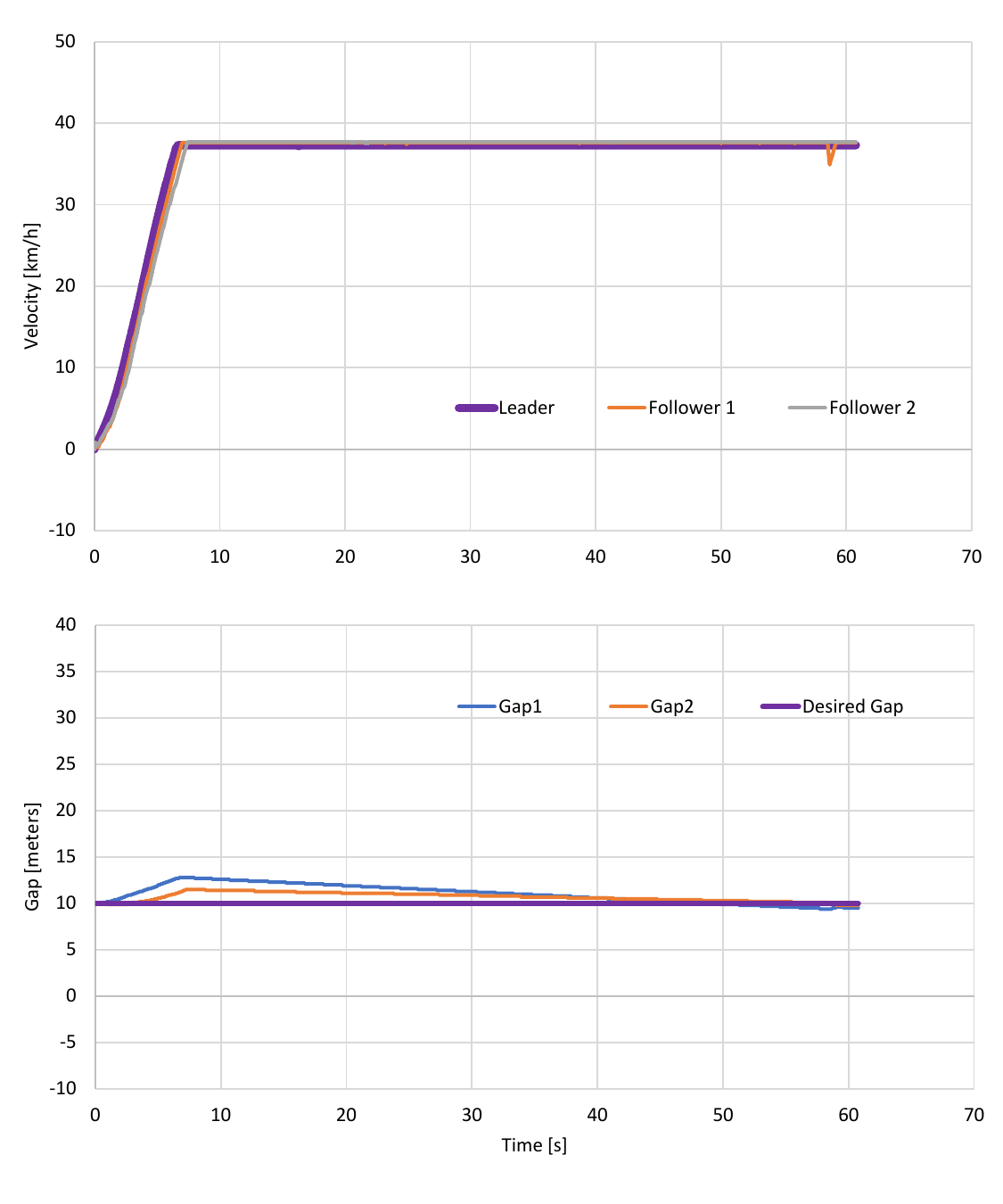}
    \caption{Gap Error and Velocity Tracking of the 3 vehicles in the platoon}
\end{figure}

\begin{figure}
    \centering
    \includegraphics[width=1\linewidth]{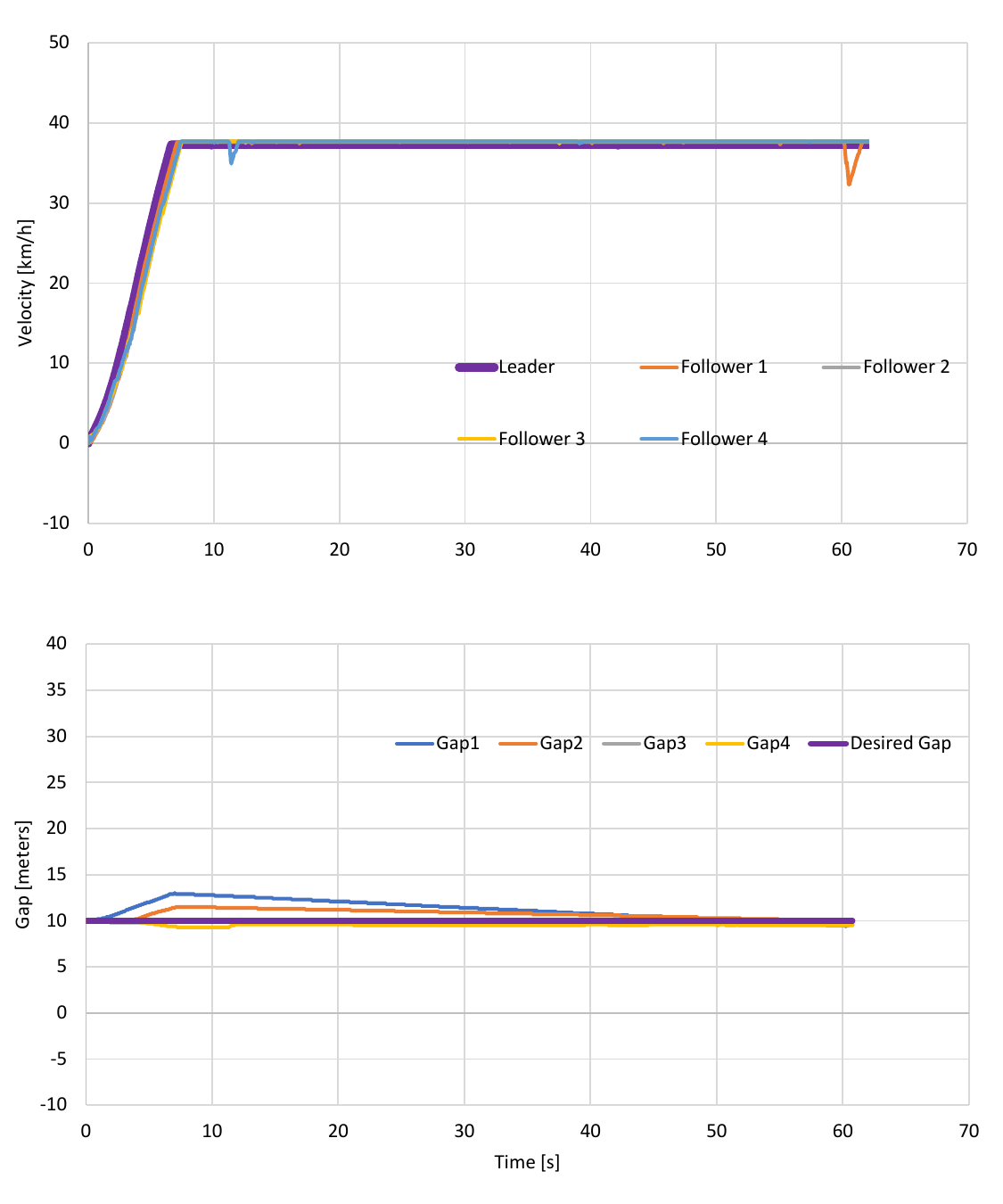}
    \caption{Gap Error and Velocity Tracking of the 5 vehicles in the platoon}
\end{figure}

Five tests with different scenarios are conducted to evaluate SDCP performance.
The first three test cases are conducted on a track of a straightforward road with no obstacles and the road turns to be able to focus more on the evaluation of the core principles of SDCP which are scalable platoon length, decentralization, and cooperation through the proposed sharing and caring communication topology. The fourth and fifth tests are conducted on the designed test track which is full of turns with tight radius of curvature and obstacles. 

\subsection{Test Scenario 1 $|$ Scalability}
\subsubsection{Pre-Testing Results}
The first test is to evaluate how the platoon can handle increasing its length from just three cars to five vehicles and then to eight connected cars, one leader, and seven followers.
Figures 11 and 12 show the gap error tracking between vehicles in the case of three vehicles and five vehicles, as a test case for scalability, respectively. After increasing the number of vehicles in the platoon to five vehicles, the gap errors between vehicles were not affected heavily and they focused on minimizing the gap error between them to minimal values as shown in the below tables(III, IV).

\begin{table}
\centering
\caption{Performance metrics for the three vehicles platoon} 
\resizebox{8.5cm}{!}{
\renewcommand{\arraystretch}{1.5}
\begin{tabular} {|c||c|c|c|c|} 
\hline
\textbf{Evaluation metrics} & \textbf{$RMSE_{Gap1}$} & \textbf{$RMSE_{Gap2}$} & \textbf{$Std_{Gap1}$} & \textbf{$Std_{Gap2}$}\\
\hline
Value (meters) & 1.45 & 0.84 & 1.01 & 0.46\\ 
\hline
\textbf{Evaluation metrics} & \textbf{$Max_{Gap1}$} & \textbf{$Max_{Gap2}$}& \textbf{$Stall_{Gap1}$} & \textbf{$Stall_{Gap2}$}\\
\hline
Value (meters) & 2.8 & 1.5 & 0 & 0.3\\ 
\hline
\end{tabular} }
\end{table}

\begin{table}
\centering
\caption{Performance metrics for the five vehicles platoon} 
\resizebox{8.5cm}{!}{
\renewcommand{\arraystretch}{1.5}
\begin{tabular} {|c||c|c|c|c|} 
 \hline
 \textbf{Evaluation metrics} & \textbf{$RMSE_{Gap1}$} & \textbf{$RMSE_{Gap2}$} & \textbf{$RMSE_{Gap3}$} & \textbf{$RMSE_{Gap4}$}\\
\hline
Value (meters) & 1.56 & 0.86 & 0.09 & 0.47\\ 
\hline
\textbf{Evaluation metrics} & \textbf{$Std_{Gap1}$} & \textbf{$Std_{Gap2}$} & \textbf{$Std_{Gap3}$} & \textbf{$Std_{Gap4}$}\\
\hline
Value (meters) & 1.01 & 0.47 & 0.08 & 0.13\\ 
\hline
\textbf{Evaluation metrics} & \textbf{$Max_{Gap1}$} & \textbf{$Max_{Gap2}$} & \textbf{$Max_{Gap3}$} & \textbf{$Max_{Gap4}$}\\
\hline
Value (meters) & 3 & 1.5 & -0.2 & -0.7\\ 
\hline
\textbf{Evaluation metrics} & \textbf{$Stall_{Gap1}$} & \textbf{$Stall_{Gap2}$} & \textbf{$Stall_{Gap3}$} & \textbf{$Stall_{Gap4}$}\\
\hline
Value (meters) & 0 & 0.2 & 0.1 & -0.5\\ 
\hline
\end{tabular} }
\end{table}

\subsubsection{Scale up platoon length to eight vehicles}
From the gap and velocity curves in Figure 13, it is observed that follower vehicles wait till the leader starts driving and then start accelerating to minimize the gap error between the leader and the first two followers while gaps between other followers are almost within the desired gap of 10 meters as shown in Table V. Tracking vehicles' velocities also assure followers' actions of accelerating to keep up with the platoon formation and maintain string stability within the platoon, and to the best of our knowledge, no work has developed such algorithm that supports stable scalability to the number of vehicles in the platoon.

\begin{table}[H]
\centering
\caption{Performance metrics of the test scenario 1 (Eight vehicles platoon)} 
\resizebox{8.55cm}{!}{
\renewcommand{\arraystretch}{2.25}
\Large
\begin{tabular} {|c||c|c|c|c|c|c|c|} 
 \hline
 \textbf{Evaluation metrics} & \textbf{$RMSE_{Gap1}$} & \textbf{$RMSE_{Gap2}$} & \textbf{$RMSE_{Gap3}$} & \textbf{$RMSE_{Gap4}$} & \textbf{$RMSE_{Gap5}$} & \textbf{$RMSE_{Gap6}$} & \textbf{$RMSE_{Gap7}$}\\
\hline
Value (meters) & 1.66 & 1.07 & 0.37 & 0.65 & 0.45 & 0.38 & 0.34\\ 
\hline
\textbf{Evaluation metrics} & \textbf{$Std_{Gap1}$} & \textbf{$Std_{Gap2}$} & \textbf{$Std_{Gap3}$} & \textbf{$Std_{Gap4}$} & \textbf{$Std_{Gap5}$} & \textbf{$Std_{Gap6}$} & \textbf{$Std_{Gap7}$}\\
\hline
Value (meters) & 1.02 & 0.64 & 0.07 & 0.25 & 0.36 & 0.26 & 0.17\\ 
\hline
\textbf{Evaluation metrics} & \textbf{$Max_{Gap5}$} & \textbf{$Max_{Gap6}$} & \textbf{$Max_{Gap5}$} & \textbf{$Max_{Gap6}$} & \textbf{$Max_{Gap5}$} & \textbf{$Max_{Gap6}$} & \textbf{$Max_{Gap7}$}\\
\hline
Value (meters) & 3.1 & 2 & -0.4 & 0.7 & 0.1 & -0.4 & 0.1\\ 
\hline
\textbf{Evaluation metrics} & \textbf{$Stall_{Gap1}$} & \textbf{$Stall_{Gap2}$} & \textbf{$Stall_{Gap3}$} & \textbf{$Stall_{Gap4}$} & \textbf{$Stall_{Gap5}$} & \textbf{$Stall_{Gap6}$} & \textbf{$Stall_{Gap7}$}\\
\hline
Value (meters) & 0 & 0 & -0.4 & 0.7 & 0.1 & -0.4 & 0.1\\
\hline
\end{tabular} }
\end{table}

\begin{figure}
    \centering
    \includegraphics[width=1\linewidth]{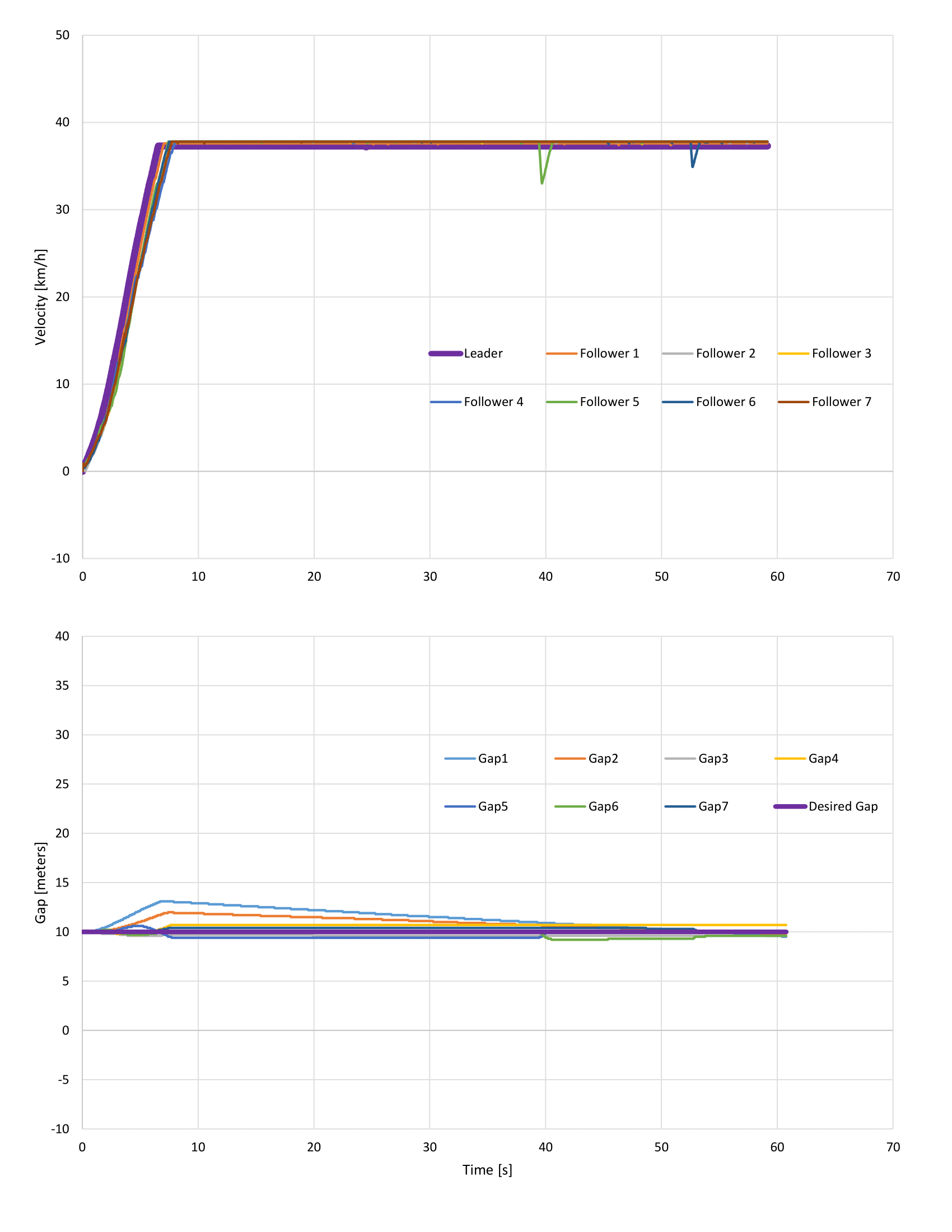}
    \caption{Gap and Velocity Tracking of Test Case 1 (eight vehicles platoon)}
\end{figure}

\subsection{Test Scenario 2 $|$ Cooperation}
The platoon's leader is subjected to aggressive behavior like a sudden change in its velocity to test the influence and coherence of the proposed communication topology on the cooperation between the vehicles in the platoon on mainly maintaining safe driving in dangerous scenarios. Figure 14 shows that vehicles behave in the first period as in test 1 accelerating to reach the desired gap between all vehicles, and then the leader's velocity suddenly decreases and keeps its speed for almost 10 seconds and then rapidly increases to its previous velocity. The platoon managed to adapt to the aggressive behavior from the leader with no crashes and decrease the gap errors between them, as shown in Table VI, but some gaps stalled with margins from the desired gap of 10 meters because of the adopted flexible spacing policy and all vehicles tend to drive at the max velocity to reach the finish line as soon as possible and collect more rewards.

\begin{table}
\centering
\caption{Performance metrics of the test scenario 2} 
\resizebox{8.55cm}{!}{
\renewcommand{\arraystretch}{2.25}
\Large
\begin{tabular} {|c||c|c|c|c|c|c|c|} 
 \hline
 \textbf{Evaluation metrics} & \textbf{$RMSE_{Gap1}$} & \textbf{$RMSE_{Gap2}$} & \textbf{$RMSE_{Gap3}$} & \textbf{$RMSE_{Gap4}$} & \textbf{$RMSE_{Gap5}$} & \textbf{$RMSE_{Gap6}$} & \textbf{$RMSE_{Gap7}$}\\
\hline
Value (meters) & 1.89 & 2.27 & 2.16 & 1.73 & 0.91 & 0.49 & 0.13\\ 
\hline
\textbf{Evaluation metrics} & \textbf{$Std_{Gap1}$} & \textbf{$Std_{Gap2}$} & \textbf{$Std_{Gap3}$} & \textbf{$Std_{Gap4}$} & \textbf{$Std_{Gap5}$} & \textbf{$Std_{Gap6}$} & \textbf{$Std_{Gap7}$}\\
\hline
Value (meters) & 1.72 & 2.17 & 1.60 & 1.67 & 0.60 & 0.30 & 0.12\\ 
\hline
\textbf{Evaluation metrics} & \textbf{$Max_{Gap5}$} & \textbf{$Max_{Gap6}$} & \textbf{$Max_{Gap5}$} & \textbf{$Max_{Gap6}$} & \textbf{$Max_{Gap5}$} & \textbf{$Max_{Gap6}$} & \textbf{$Max_{Gap7}$}\\
\hline
Value (meters) & 3.2 & -5.6 & -6 & -4.7 & -2.7 & -0.9 & -0.3\\ 
\hline
\textbf{Evaluation metrics} & \textbf{$Stall_{Gap1}$} & \textbf{$Stall_{Gap2}$} & \textbf{$Stall_{Gap3}$} & \textbf{$Stall_{Gap4}$} & \textbf{$Stall_{Gap5}$} & \textbf{$Stall_{Gap6}$} & \textbf{$Stall_{Gap7}$}\\
\hline
Value (meters) & -0.8 & -3.4 & -1.9 & -3.9 & -2.7 & -0.6 & -0.3\\
\hline
\end{tabular} }
\end{table}

\begin{figure}
    \centering
    \includegraphics[width=1\linewidth]{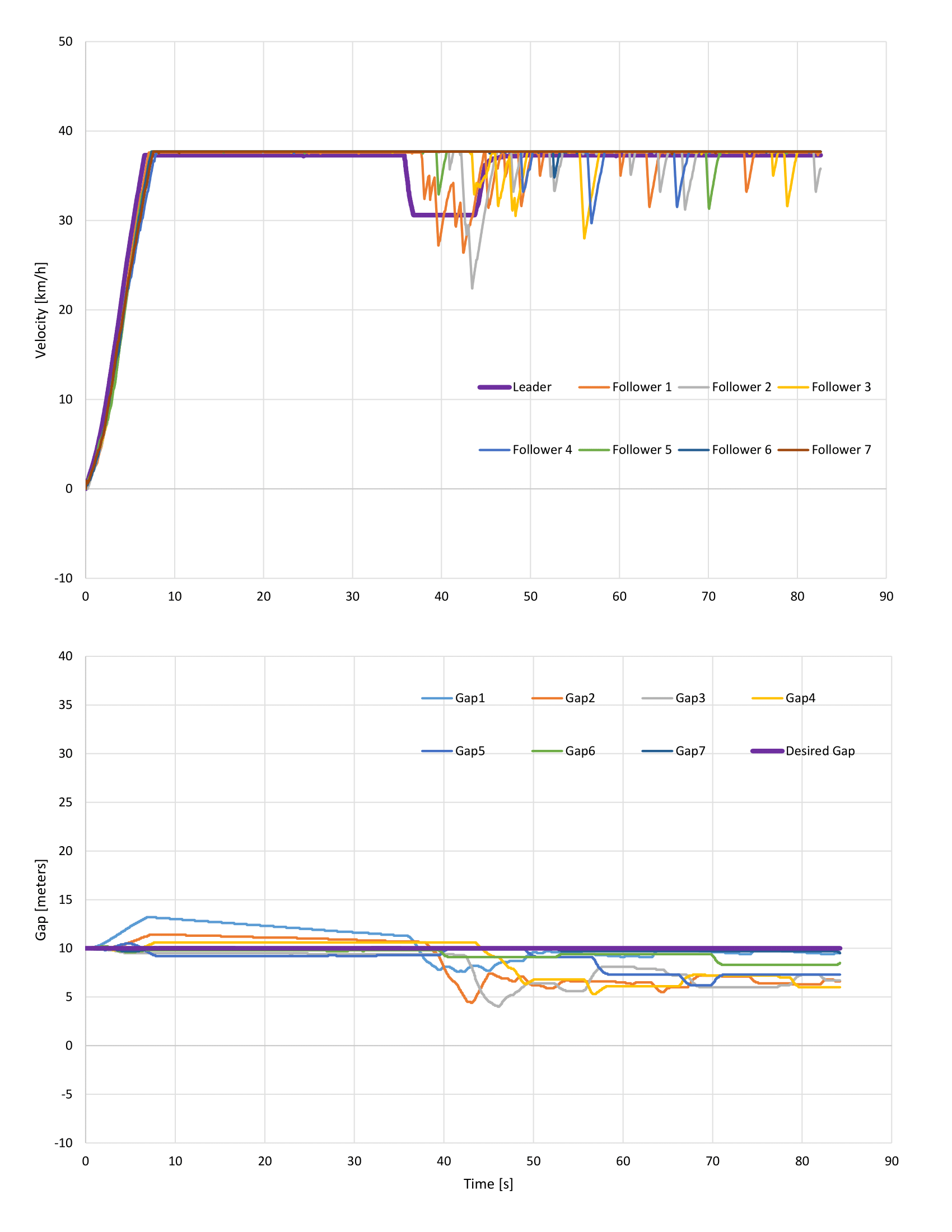}
    \caption{Gap and Velocity Tracking of Test Case 2}
\end{figure}

\subsection{Test Scenario 3 $|$ Decentralization}
In this test, followers will be ordered to drive without the leader's vehicle and their performance is measured to be a clear validation of the decentralized behavior of the vehicles in the platoon. followers significantly are able to drive without the leader and continue depending on each other while sharing their perception data between them through V2V, and almost all of them maintained the desired gap with minimal errors, as shown in Table VII, and the same velocity between them as shown in figure 15. This is proof of one of the advantages of having a decentralized system for such cases the vehicles can handle emergency cases when the leader of the platoon, where followers are used to driving following it, is not able to drive and they should adapt to this new and unexpected circumstances.

\begin{table}
\centering
\caption{Performance metrics of the test scenario 3} 
\resizebox{8.55cm}{!}{
\renewcommand{\arraystretch}{2.25}
\Large
\begin{tabular} {|c||c|c|c|c|c|c|c|} 
 \hline
 \textbf{Evaluation metrics} & \textbf{$RMSE_{Gap1}$} & \textbf{$RMSE_{Gap2}$} & \textbf{$RMSE_{Gap3}$} & \textbf{$RMSE_{Gap4}$} & \textbf{$RMSE_{Gap5}$} & \textbf{$RMSE_{Gap6}$} & \textbf{$RMSE_{Gap7}$}\\
\hline
Value (meters) & 347.22 & 0.32 & 1.11 & 0.55 & 0.49 & 0.84 & 0.64\\ 
\hline
\textbf{Evaluation metrics} & \textbf{$Std_{Gap1}$} & \textbf{$Std_{Gap2}$} & \textbf{$Std_{Gap3}$} & \textbf{$Std_{Gap4}$} & \textbf{$Std_{Gap5}$} & \textbf{$Std_{Gap6}$} & \textbf{$Std_{Gap7}$}\\
\hline
Value (meters) & 141.26 & 0.30 & 0.37 & 0.20 & 0.18 & 0.28 & 0.15\\ 
\hline
\textbf{Evaluation metrics} & \textbf{$Max_{Gap5}$} & \textbf{$Max_{Gap6}$} & \textbf{$Max_{Gap5}$} & \textbf{$Max_{Gap6}$} & \textbf{$Max_{Gap5}$} & \textbf{$Max_{Gap6}$} & \textbf{$Max_{Gap7}$}\\
\hline
Value (meters) & 441.8 & 1.6 & -1.5 & -1 & 0.6 & -1 & -0.8\\ 
\hline
\textbf{Evaluation metrics} & \textbf{$Stall_{Gap1}$} & \textbf{$Stall_{Gap2}$} & \textbf{$Stall_{Gap3}$} & \textbf{$Stall_{Gap4}$} & \textbf{$Stall_{Gap5}$} & \textbf{$Stall_{Gap6}$} & \textbf{$Stall_{Gap7}$}\\
\hline
Value (meters) & 441.8 & 0.2 & -0.6 & -0.9 & 0.5 & -0.7 & -0.8\\
\hline
\end{tabular} }
\end{table}

\begin{figure}
    \centering
    \includegraphics[width=1\linewidth]{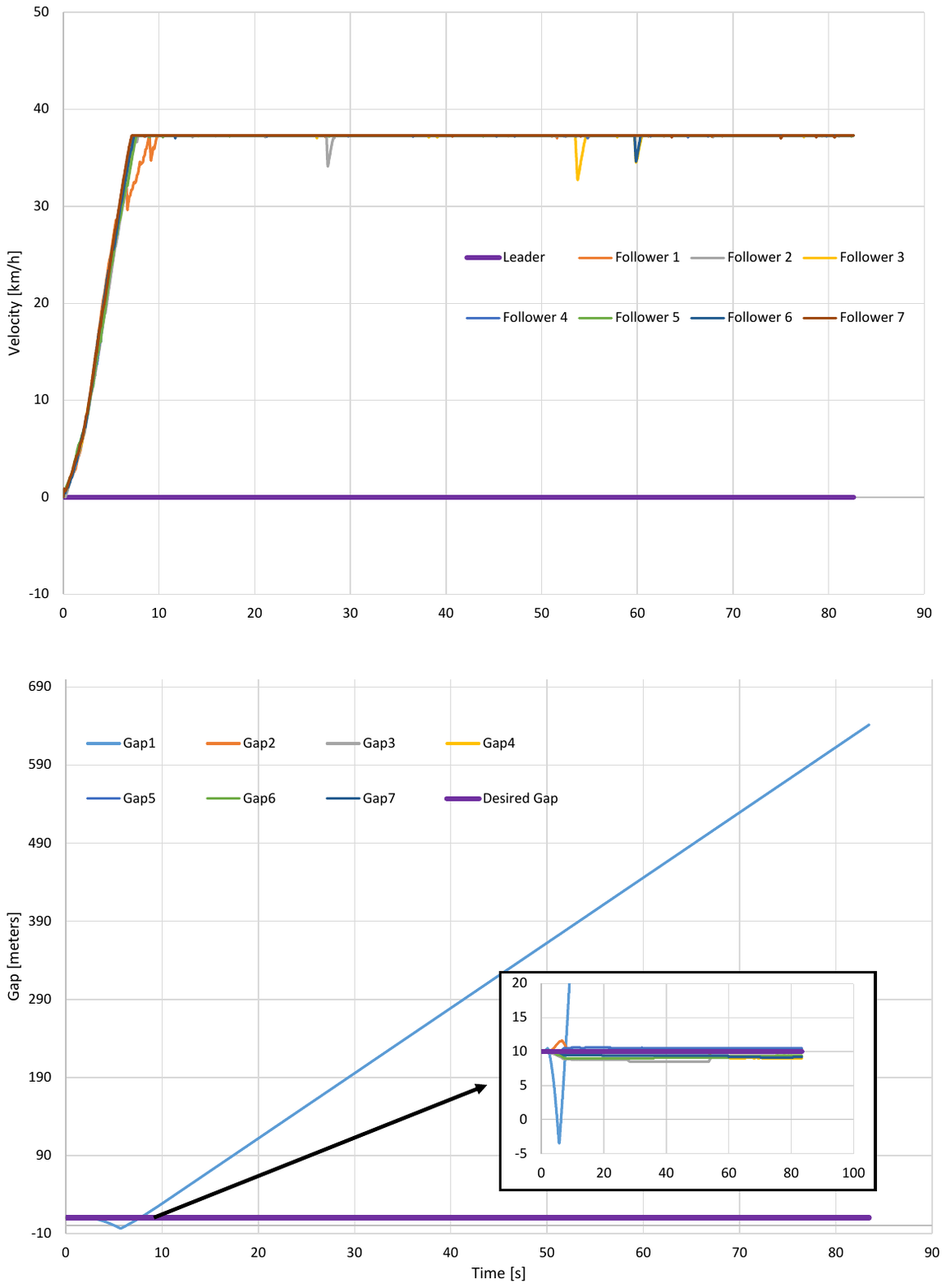}
    \caption{Gap and Velocity Tracking of Test Case 3}
\end{figure}

\subsection{Test Scenario 4 $|$ SDCP on the urban testing track}
Figure 16 shows the gaps and velocities of the platoon's vehicles driving on the test track and how followers try to maintain the desired gap with small margins, as shown in Table VIII, and the same velocity through all the taken maneuvers to drive safely through the consecutive tight turns, obstacles, and experience a new scenario of passing through a narrow way between two obstacles. In such a hard environment, vehicles give higher priority to ego performance of keeping driving safely through the whole track, collecting more positive rewards from reaching more checkpoints and reaching the finish line to gain the highest reward and avoid high negative rewards caused by crashing, than to focus only on maintaining the desired gap. Decentralized driving behavior and flexible spacing policy effects are observed in the performance of the vehicles as each vehicle has its own objectives and behavior that respect the platoon's regulations and formation while focusing on solo driving performance. Based on this acquired behavior and driving skills through training using deep RL, vehicles achieve both individual stability and string stability.

\begin{table}
\centering
\caption{Performance metrics of the test scenario 4} 
\resizebox{8.55cm}{!}{
\renewcommand{\arraystretch}{2.25}
\Large
\begin{tabular} {|c||c|c|c|c|c|c|c|} 
 \hline
 \textbf{Evaluation metrics} & \textbf{$RMSE_{Gap1}$} & \textbf{$RMSE_{Gap2}$} & \textbf{$RMSE_{Gap3}$} & \textbf{$RMSE_{Gap4}$} & \textbf{$RMSE_{Gap5}$} & \textbf{$RMSE_{Gap6}$} & \textbf{$RMSE_{Gap7}$}\\
\hline
Value (meters) & 2.49 & 0.71 & 1.94 & 2.34 & 0.53 & 1.27 & 2.03\\ 
\hline
\textbf{Evaluation metrics} & \textbf{$Std_{Gap1}$} & \textbf{$Std_{Gap2}$} & \textbf{$Std_{Gap3}$} & \textbf{$Std_{Gap4}$} & \textbf{$Std_{Gap5}$} & \textbf{$Std_{Gap6}$} & \textbf{$Std_{Gap7}$}\\
\hline
Value (meters) & 0.79 & 0.59 & 1.25 & 1.41 & 0.46 & 1.01 & 1.53\\ 
\hline
\textbf{Evaluation metrics} & \textbf{$Max_{Gap5}$} & \textbf{$Max_{Gap6}$} & \textbf{$Max_{Gap5}$} & \textbf{$Max_{Gap6}$} & \textbf{$Max_{Gap5}$} & \textbf{$Max_{Gap6}$} & \textbf{$Max_{Gap7}$}\\
\hline
Value (meters) & 4.3 & -2.6 & 3.2 & 3.9 & -2.6 & -3.5 & -3.9\\ 
\hline
\textbf{Evaluation metrics} & \textbf{$Stall_{Gap1}$} & \textbf{$Stall_{Gap2}$} & \textbf{$Stall_{Gap3}$} & \textbf{$Stall_{Gap4}$} & \textbf{$Stall_{Gap5}$} & \textbf{$Stall_{Gap6}$} & \textbf{$Stall_{Gap7}$}\\
\hline
Value (meters) & 1.8 & 0.1 & 1 & 3.8 & -0.3 & -1.4 & 1.6\\
\hline
\end{tabular} }
\end{table}

\begin{figure}
    \centering
    \includegraphics[width=1\linewidth]{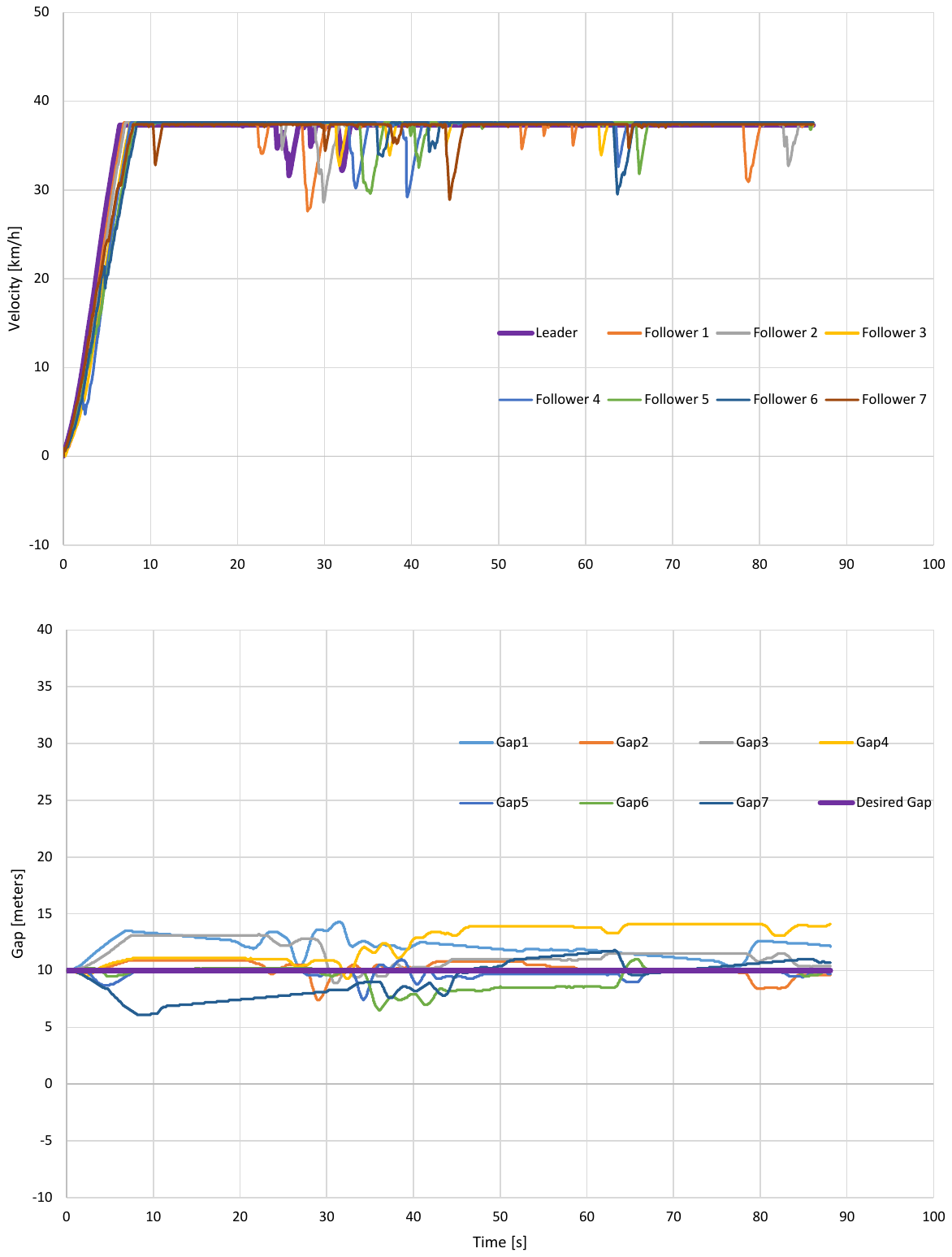}
    \caption{Gap and Velocity Tracking of Test Case 4}
\end{figure}

\subsection{Test Scenario 5 $|$ Traffic Congestion}
A platoon with eight vehicles was subjected to drive on the testing track, as shown in Figure 9 while being connected based on predecessor follower-based sharing and caring communication topology and without being connected to measure how the communication and cooperation between vehicles affect traffic congestion and safe driving. Results show that one accident occurred between the last two vehicles in the platoon when there was no communication applied between vehicles, and with communication, no accidents occurred and the throughput increased by 33\%.

\begin{table}[H]
\centering
\caption{Number of Accidents for each test case} 
\resizebox{8.55cm}{!}{
\renewcommand{\arraystretch}{2.25}
\Large
\begin{tabular} {|c||c|c|c|c|c|c|} 
 \hline
 \textbf{Test Case no.} & \textbf{Test Case 1} & \textbf{Test Case 2} & \textbf{Test Case 3} & \textbf{Test Case 4} & \textbf{Test Case 5 (V2V on)} & \textbf{Test Case 5 (V2V off)}\\
\hline
 Accidents & 0 & 0 & 0 & 0 & 0 & 1\\
\hline
\end{tabular} }
\end{table}

\subsection{Discussion}
Numerical results in the previous tables (III-IX) show the metrics of performance of all the gap errors, the velocities tracking of the vehicles in the platoon, and the number of crashes for each test case study through the whole driving distance.\\
SDCP completely succeeded in reaching zero crashes in all the case studies, and this boosts the gained behavior that gives high priority to safe driving through the applied sharing and caring communication topology and the reward signals that give high penalties on crashes. RMSE, standard deviation, and stalled gap errors for all gaps are relatively small compared to the total driving distances and based on the influence of the adopted flexible spacing policy which gives more driving flexibility for the vehicles within the platoon, as long as no accidents occur, except for the first follower in test scenario 3 which gets a very high penalty on the gap error between it and the leader, predecessor vehicle, but since follower's behavior is decentralized and has more courage to keep drive safely than to stop and wait for the leader to lead the platoon, the first follower continued driving as if it is the leader and followers cooperatively followed each other till reached the finish line and all gain the highest positive rewards in the environment.\\
Decentralization improves the vehicle's solo performance, but it affects the synchronization between vehicles and this is one of the limitations of decentralized systems. Vehicles' behavior in some situations, was greedy for their performance and cumulative reward, however, on the other hand, decentralization with the help of the novel predecessor follower-based sharing and caring communication topology made it applicable to have a robust scalable platoon, guaranteeing platoon stability, and safe driving by achieving zero accidents through the five conducted test case seniors (excluding when V2V is off in test case 5).

\subsection{Work Limitations}
There are some limitations in the conducted work listed as follows:
\begin{itemize}
    \item Roads in simulation work must have borders since only depth sensors are used.
    \item All simulation work is conducted on roads with only 2 lanes.
    \item Testing track created in Unity can only handle the presence of a platoon with a maximum of 8 vehicles and the track must get bigger to handle more vehicles for stable empirical experiments.
\end{itemize}

%% file: Sections/sec6_conclusion.tex
\section{Conclusion}
In this paper vehicle platooning is developed by targeting three main aspects of scalable platoon length, decentralization, and positive cooperative driving. A novel predecessor follower-based sharing and caring communication topology are proposed, which is based on sharing perception data between the connected vehicles and caring by sending positive or negative rewards indicating positive or negative cooperation occurred. The vehicle platooning model is trained using deep RL on Unity 3D game engine which has quite good physics that mimics reality and with ml-agents toolkit that provides the ability to build complex 3D environments and train intelligent agents in them all while leveraging the powerful Unity engine and user interface. Vehicles are trained on a track with tight turns and obstacles to simulate tough maneuvers that are hard for humans themselves. Results show how empirically the developed SDCP performed positively on the five proposed test case scenarios and increased throughput by 33\%. Vehicles were able to drive safely reaching zero accidents with very low RMSE, standard deviation, and stalled gap errors of the gap errors between vehicles in the platoon relative to total driving distances. Also, they were able to handle the leader's aggressive behavior or complex road scenarios like passing through a narrow way between two obstacles, and even driving without a leader although all these test cases are new scenarios for the trained model of the developed SDCP and vehicles have no prior knowledge about the environment or the track design, and this proofs the robustness of SDCP against various hard scenarios.\\
Some of the recommendations to improve this research work are to study various control algorithms on vehicle platooning motion and formation control to enhance platoon performance on handling both safe solo driving and cooperative driving simultaneously, and use a camera for visual perception and fuse it with the depth sensors readings to have a better perception of the surrounding environment.

%% file: Sections/ref.tex
\bibliographystyle{IEEEtran}